\documentclass[12pt]{article}
\pdfoutput=1
\usepackage{graphicx,pict2e,latexsym,xcolor}
\usepackage[square,numbers]{natbib}
\newtheorem{defn}{Definition}
\newtheorem{theorem}{Theorem}
\newtheorem{lemma}{Lemma}

\newtheorem{example}{Example}

\def\M{{\mathcal M}}
\def\G{{\mathcal G}}
\def\B{{\mathcal B}}

\newcommand{\iid}{\stackrel{iid}{\sim}}

\begin{document}
\bibliographystyle{plainnat}
\pagestyle{plain}

\title{\Large \bf Rehabilitating Isomap: \\
Euclidean Representation of Geodesic Structure}

\author{Michael W. Trosset\thanks{Department of Statistics, Indiana University.  
E-mail: {\tt mtrosset@indiana.edu}} 
\and
G\"{o}k\c{c}en B\"{u}y\"{u}kba\c{s}\thanks{Department of Mathematics, Indiana University. E-mail: {\tt gokbuyuk@indiana.edu}}
}

\date{\today}

\maketitle


\begin{abstract}
Manifold learning techniques for nonlinear dimension reduction assume that high-dimensional feature vectors lie on a low-dimensional manifold, then attempt to exploit manifold structure to obtain useful low-dimensional Euclidean representations of the data.
Isomap, a seminal manifold learning technique, is an elegant synthesis of two simple ideas: the approximation of Riemannian distances with shortest path distances on a graph that localizes manifold structure, and the approximation of shortest path distances with Euclidean distances by multidimensional scaling.
We revisit the rationale for Isomap, clarifying what Isomap does and what it does not.  In particular, we explore the widespread perception that Isomap should only be used when the manifold is parametrized by a convex region of Euclidean space.
We argue that this perception is based on an extremely narrow interpretation of manifold learning as parametrization recovery, and we submit that Isomap is better understood as constructing Euclidean representations of geodesic structure.  We reconsider a well-known example that was previously interpreted as evidence of Isomap's limitations, and we re-examine the original analysis of Isomap's convergence properties, concluding that convexity is not required for shortest path distances to converge to Riemannian distances.
\end{abstract}


\newpage

\tableofcontents

\newpage


\section{Introduction}
\label{intro}

Multivariate data are often represented as points in an 
ambient feature space, e.g., $x_1,\ldots,x_n \in \Re^q$.
By {\em dimension reduction}, we mean the representation of
$x_1,\ldots,x_n$ as $z_1,\ldots,z_n \in \Re^d$ for $d<q$.
Principal component analysis (PCA)
constructs $z_1,\ldots,z_n$ by
projecting $x_1,\ldots,x_n$ into a $d$-dimensional hyperplane,
a classic example of a {\em linear dimension reduction}\/
technique.  Sometimes, however, one can obtain a more parsimonious 
representation of the data by performing
{\em nonlinear dimension reduction}.  For example, suppose
that we sample points in $\Re^2$ along the sine wave
$(t, \sin t)$.  
By straightening the sine wave
we obtain a perfect $1$-dimensional representation of the data.
This procedure is nonlinear: 
any projection of the data
into a straight line will distort the arc length distances
between the points,
i.e., the distances measured along the trajectory of the sine wave. 

A sine wave is an example of a $1$-dimensional manifold,
i.e., its local structure resembles $\Re$.
The phrase {\em manifold learning}\/ encompasses a variety of techniques for nonlinear dimension reduction,
each motivated by the conceit that $x_1,\dots,x_n$ lie on (or near) a low-dimensional manifold in $\Re^q$.  
Do actual multivariate data lie 
(approximately) on low-dimensional manifolds?
Describing ``the neglected case of nonlinear data structures,''
Shepard and Carroll \cite{shepard&carroll:1966} argued that
\begin{quote}
``there may well be strong nonlinear relations among the variables.
If so, the objects will not scatter in all directions according, say, to some ellipsoidal distribution in the multivariate space.  Instead, they will tend to fall on some manifold, of lower intrinsic dimensionality, that may nevertheless curve and twist through the space in such a way as to give the superficial {\em appearance}\/ of filling an ellipsoidal volume.''
\end{quote}
More recently, Roweis and Saul \cite{lle:2000} argued that  
``Coherent structure in the world leads to strong correlations between inputs{\ldots}, generating observations that lie on or close to a smooth low-dimensional manifold.''
Manifold learning is concerned with such situations.

The present investigation revisits Isomap \cite{isomap:2000},
a seminal manifold learning technique.  Despite---or conceivably because of---its simplicity, Isomap has declined in popularity since its introduction in 2000.  We endeavor to understand why, and to correct some misconceptions that are widely associated with the technique.  Section~\ref{prelim} collects some relevant background on manifolds, Riemannian geometry, Euclidean distance geometry, and multidimensional scaling.  Section~\ref{motivation} describes several attempts to construct Euclidean representations  of non-Euclidean manifolds.  Section~\ref{isomap} describes Isomap and some ambiguities that confound its use.  Section~\ref{recovery} discusses the Parametrization Recovery Problem, one possible way of stating what Isomap is supposed to accomplish.  Section~\ref{converge} revisits the convergence analysis \cite{Bernstein&etal:2000} that accompanied the introduction of Isomap.  Section~\ref{discuss} concludes.

\section{Mathematical Preliminaries}
\label{prelim}

This section collects various mathematical definitions and results that are essential to our exposition of Isomap.

\subsection{Manifolds}

The following definition appears in \cite{milnor:1965}.

\begin{defn} 
A set ${\mathcal M} \subset \Re^q$ 
is called a smooth manifold of dimension $p$ 
if and only if each $m \in {\mathcal M}$ has a neighborhood 
that is diffeomorphic to an open subset of $\Re^p$.
\label{def:manifold}
\end{defn}
\medskip
Several elements of this definition require elaboration:
First, in this context, a {\em neighborhood}\/ of $m$ is the intersection of ${\mathcal M}$ and an open set $W \subset \Re^q$.
Second, the sets $W \cap {\mathcal M}$ and $U \subset \Re^p$ are {\em diffeomorphic}\/ if there is a one-to-one function $g : U \rightarrow W \cap {\mathcal M}$ such that both $g$ and $g^{-1}$ are smooth.  The function $g$ is a {\em parametrization}\/ of $W \cap {\mathcal M}$,
whereas the function $g^{-1}$ induces a system of {\em coordinates}\/ on $W \cap {\mathcal M}$.
Third, a function is $C^r$ if its derivatives of order $r$ are continuous.
We may understand {\em smooth}\/ to specify a specific order of differentiability (e.g., $r=1$ in \cite{milnor:1965}), or in the somewhat more vague sense of ``as many derivatives as the situation requires.''  In Definition~\ref{def:manifold},
the smoothness of ${\mathcal M}$ is determined by the smoothness of $g$.
Fourth, the dimension $p$ is fixed, i.e., it may not vary with $m$.

Here are some elementary examples of low-dimensional manifolds.

\begin{example}[Spirals]
Informally, a plane spiral is a smooth curve in $\Re^2$ that winds around a fixed central point $m$ at a monotonically increasing distance from $m$.  For example, represent $\Re^2$ in polar coordinates $(r,\theta)$ with origin $m$. 
Then $\sigma(t) = (\beta t,t)$, for $\beta>0$, is an Archimedean spiral.
If $[a,b] \subset (0,\infty)$, then $\sigma([a,b])$ is a compact connected $1$-dimensional manifold embedded in a $2$-dimensional ambient space.  
\label{ex:spiral}
\end{example}

\begin{example}[Swiss Rolls]
According to Wikipedia, a {\em Swiss roll}\/ is a rolled cake spread with jelly (or jam, whipped cream, icing, etc.).  Its spiral shape suggests an obvious extension of Example~\ref{ex:spiral}.  Suppose that $\sigma : [a,b] \rightarrow \Re^2$ parametrizes a spiral.  Then $\tau : [a,b] \times [c,d] \rightarrow \Re^3$ defined by
$\tau(s,t) = (\sigma(s),t)$ parametrizes the mathematical abstraction of a Swiss roll, a compact connected $2$-dimensional manifold embedded in a $3$-dimensional ambient space.

Swiss rolls have played an outsized role in the brief history of manifold learning.
Figure~3 in \cite{isomap:2000} used $1000$ points on a Swiss roll to demonstrate Isomap, 
Figure~1 in \cite{lle:2000} used points on a Swiss roll to demonstrate {\em Locally Linear Embedding},
and numerous other researchers have followed suit.  In Section~\ref{recovery} we will suggest a reason why Swiss rolls pervade the manifold learning literature.
\label{ex:swiss}
\end{example}

\begin{example}[Hemispheres]
A great circle divides the sphere in which it resides into two opposing halves, each of which is a {\em hemisphere}.  For example, the unit sphere in $\Re^3$ is
\[
S^2 = \left\{ \left( x_1,x_2,x_3 \right) \in \Re^3 \; : \; x_1^2+x_2^2+x_3^2=1 \right\}
\]
and the northern hemisphere of $S^2$ is the set of points
\[
\left\{ \left( x_1,x_2,x_3 \right) \in S^2 \; : \; x_3 \geq 0 \right\},
\]
another compact connected $2$-dimensional manifold embedded in a $3$-dimensional ambient space.
\label{ex:hemi}
\end{example}

\subsection{Riemannian Geometry}

A {\em metric tensor}\/ on the manifold ${\mathcal M}$ is a collection of inner products on the tangent spaces of ${\mathcal M}$.  If ${\mathcal M}$ admits a metric tensor, then ${\mathcal M}$ is a {\em Riemannian manifold}.  See \citep[Part II]{milnor:1963} for a rapid course in Riemannian geometry and \citep{Lee:1997} for a more expansive development.  Note that many authors refer to the metric tensor as a {\em Riemannian metric}.  In neither expression is the word ``metric'' used in the sense of a distance function.

For $a<b$, the smooth curve $\gamma: [a,b] \rightarrow {\mathcal M}$ has length 
\[
L(\gamma) = \int_a^b \left\| \dot{\gamma}(t) \right\| \; dt.
\]
It is {\em parametrized by arc length}\/ if and only if $\| \dot{\gamma}(t) \| =1$ for every $t \in (a,b)$, in which case the length of $\gamma: [a,t] \rightarrow {\mathcal M}$ is $t-a$.
It is a {\em geodesic}\/ if and only if $\| \ddot{\gamma}(t) \| =0$ for every $t \in (a,b)$.  The concept of a geodesic curve extends the concept of a straight line from Euclidean space to Riemannian manifolds.

Suppose that ${\mathcal M}$ is connected.  The {\em Riemannian distance}\/ between $m_1,m_2 \in {\mathcal M}$, $d_{\mathcal M}(m_1,m_2)$, is the infimum of the lengths of all curves with endpoints $m_1$ and $m_2$.  If $\gamma(a)=m_1$, $\gamma(b)=m_2$, and $L(\gamma)=d_{\mathcal M}(m_1,m_2)$, then $\gamma$ is {\em minimizing}.
When parametrized by arc length, every minimizing curve is a geodesic \citep[Theorem~6.6]{Lee:1997}.  This result extends the familiar fact that ``the shortest distance between two points is a straight line'' from Euclidean geometry to Riemannian geometry.
Conversely, every geodesic curve in ${\mathcal M}$ is locally minimizing \citep[Theorem~6.12]{Lee:1997}.  Furthermore,
if ${\mathcal M}$ is also compact, then it follows from the celebrated Hopf--Rinow Theorem that any two points in ${\mathcal M}$ can be joined by a minimizing geodesic curve \citep[Corollary~6.16, Theorem~6.13, and Corollary~6.15]{Lee:1997}.

Riemannian distance induces a metric topology on $\M$, and the metric topology is equivalent to the topology induced by the definition of manifold \cite[Lemma~6.2]{Lee:1997}.  Let $\B$ denote the Borel sigma-field on $\M$, i.e., the smallest sigma-field that contains the open sets in $\M$.  Then $(\M,\B)$ is a measurable space, allowing the construction of probability measures from which samples of points in $\M$ can be drawn.

An {\em isometry}\/ between two metric spaces is a smooth distance-preserving map from one to the other.  An isometry is necessarily injective; if it is also bijective, then it is a {\em global isometry}.  Two metric spaces are {\em globally isometric}\/ if there exists a global isometry between them.  A metric space $\M_1$ is {\em locally isometric}\/ to a metric space $M_2$ if each point in $\M_1$ has a neighborhood that is globally isometric to an open set in $\M_2$.

\subsection{Euclidean Distance Geometry}

Given an $n \times n$ matrix $\Delta = [ \delta_{ij} ]$,
the fundamental problem of Euclidean distance geometry is
to determine whether or not there exist $p$ and
$z_1,\ldots,z_n \in \Re^p$ such that each
$\delta_{ij} = \| z_i-z_j \|$.
If such a configuration of points exist,
then we say that $\Delta$ is a Type~1 Euclidean distance matrix
(EDM-1).  The configuration itself is an {\em embedding}\/
of $\Delta$ in $\Re^p$, and the smallest $p$ for which
embedding is possible is the 
{\em embedding dimension}\/ of $\Delta$.
If there exists a configuration such that each
$\delta_{ij} = \| z_i-z_j \|^2$, then we say that $\Delta$
is a Type~2 Euclidean distance matrix (EDM-2).  Obviously, $\Delta$ is EDM-1 if and only if $\Delta_2 = [ \delta_{ij}^2 ]$
is EDM-2.

Several easily checked conditions that are necessary for
$\Delta$ to be EDM-1 are readily inferred from the definition
of distance.  If $\Delta$ is EDM-1, then
$\delta_{ij}=\delta_{ji}$ ($\Delta$ is symmetric),
$\delta_{ij} \geq 0$ ($\Delta$ is nonnegative), and
$\delta_{ii}=0$ ($\Delta$ is hollow).
A symmetric, nonnegative, hollow matrix---a matrix
that might plausibly be EDM-1 (or might naturally be 
approximated by a matrix that is EDM-1)---is a {\em dissimilarity matrix}.

The following result provides a constructive solution to the problem of determining whether or not a dissimilarity matrix $A$ is EDM-2.  To determine if $\Delta$ is EDM-1, apply Theorem~\ref{thm:embed.e} to $A=\Delta_2$.
\begin{theorem}
Let $A= [ a_{ij} ]$ denote an $n \times n$ dissimilarity matrix.
Let $P=I-ee^t/n$, where $I$ is the $n \times n$ identity matrix and $e=(1,\ldots,1) \in \Re^n$.
Then $A$ is EDM-2 if and only if the symmetric matrix 
\[
B = \tau \left( A \right) =
-\frac{1}{2} P A P 
\]
is positive semidefinite ($B \geq 0$).
If $B=[ b_{ij} ] \geq 0$, then $\mbox{\rm rank}(B)$
is the embedding dimension of $A$.
If $z_1,\ldots,z_n \in \Re^p$ are such that
$\langle z_i,z_j \rangle = b_{ij}$,
then $\| z_i-z_j \|^2 = a_{ij}$.
\label{thm:embed.e}
\end{theorem}
The usual approach to embedding an EDM-2 matrix $A$ begins by
computing the spectral decomposition of $B \geq 0$ and
writing
$B = \sum_{i=1}^r \lambda_i u_i u_i^t$,
where $\lambda_1 \geq \cdots \geq \lambda_r >0$
are the strictly positive eigenvalues of $B$
and $u_1,\ldots,u_r$ are corresponding
orthonormal eigenvectors.
Set $\lambda_i = \sigma_i^2$;
then the $n \times r$ {\em configuration matrix}\/
\[
Z = \left[ \begin{array}{c|c|c}
\sigma_1 u_1 & \cdots & \sigma_r u_r
\end{array} \right] =
\left[ \begin{array}{c}
z_1^t \\ \vdots \\ z_n^t
\end{array} \right] 
\]
is an embedding of $\Delta$.  
Embeddings obtained from Theorem~\ref{thm:embed.e} center the configuration at the origin of $\Re^r$, a choice popularized by Torgerson \cite{torgerson:1952}.  Analogous embeddings that place the origin at $x_n$ were proposed by Schoenberg \cite{schoenberg:1935} and by Young and Householder \cite{young&householder:1938}.
The general case was considered by Gower 
\cite{gower:1982,gower:1985}.

\subsection{Multidimensional Scaling}

Multidimensional scaling (MDS) is a collection of techniques for constructing Euclidean configurations from dissimilarity matrices that may not be EDM-1.  Classical multidimensional scaling (CMDS), proposed by Torgerson \cite{torgerson:1952}, is based on Theorem~\ref{thm:embed.e}.  To construct a configuration $z_1,\ldots,z_n \in \Re^d$ from $\Delta$, first compute $B = \tau(\Delta_2)$, its $d$ largest eigenvalues $\lambda_1 \geq \cdots \geq \lambda_d$, and corresponding
orthonormal eigenvectors $u_1,\ldots,u_d$;
then set $\bar{\lambda}_i = \max(\lambda_i,0) = \sigma_i^2$, and
\[
Z = \left[ \begin{array}{c|c|c}
\sigma_1 u_1 & \cdots & \sigma_d u_d
\end{array} \right] =
\left[ \begin{array}{c}
z_1^t \\ \vdots \\ z_n^t
\end{array} \right] .
\]
The resulting configuration is centered at the origin and its Cartesian coordinate axes are its
principal component axes.

The following optimality property is implicit in \cite{torgerson:1952}; \cite{mardia:1978} contains a formal proof.
\begin{theorem}
Let
\[
B = U \Lambda U^t = \sum_{i=1}^n \lambda_i u_i u_i^t
\]
be the spectral decomposition of the $n \times n$ symmetric matrix $B$,
with eigenvalues $\lambda_1 \geq \cdots \geq \lambda_n$.
Define $\bar{\lambda}_i = \max(\lambda_i,0)$ for $i=1,\ldots,d$,
$\bar{\lambda}_i = 0$ for $i=d+1,\ldots,n$, 
$\bar{\Lambda} = \mbox{\rm diag}(\bar{\lambda}_1,\ldots,\bar{\lambda}_n)$,
and
\[
\bar{B} = U \bar{\Lambda} U^t = \sum_{i=1}^d \bar{\lambda}_i u_i u_i^t.
\]
If $C$ is any $n \times n$ symmetric positive semidefinite matrix of rank $\leq d$, then
\[
\left\| \bar{B}-B \right\|_F^2 \leq
\left\| C-B \right\|_F^2 .
\]
\label{thm:psd}
\end{theorem}
It follows that CMDS constructs the $d$-dimensional configuration whose pairwise inner products best approximate (in the sense of squared error) the ``fallible'' inner products $B =\tau(\Delta_2)$.

CMDS does not (directly) approximate dissimilarities with Euclidean distances.  To do so, one might embed $\Delta$ by choosing $Z$ to minimize Kruskal's \cite{kruskal:1964a} {\em raw stress criterion},
\begin{equation}
\sigma(Z) = \sum_{i \leftrightarrow j} w_{ij} \left[ d_{ij}(Z)-\delta_{ij} \right]^2
= \frac{1}{2} \sum_{i,j=1}^n w_{ij} \left[ \left\| z_i-z_j \right\| -\delta_{ij} \right]^2,
\label{eq:stress}
\end{equation}
where $w_{ij}=w_{ji} \geq 0$ is the weight assigned to approximating
$\delta_{ij}=\delta_{ji}$ with $\| z_i-z_j \| = \| z_j-z_i \|$.
The configuration that minimizes the raw stress criterion is,
in the sense of weighted squared error, the configuration whose
interpoint distances best approximate the specified dissimilarities.
The incorporation of weights into the error criterion provides enormous flexibility.

In contrast to CMDS, minimizing the raw stress criterion requires numerical optimization.  
This is often accomplished by repeated iterations of the {\em Guttman transformation}, described in \cite[Chapter 8]{borg&groenen:2005}.  At least when $w_{ij}=1$ and the configuration is initialized by CMDS, several iterations usually result in a nearly optimal embedding.  If desired, Newton's method 
\citep{kearsley&etal:newton} can be used to obtain more precise solutions.  If $n$ is large enough that CMDS is prohibitively expensive, then one can construct a less expensive initial configuration by {\em landmark multidimensional scaling}\/ 
\cite{desilva&tenenbaum:2003,desilva&tenenbaum:2004}.

\section{Motivating Examples}
\label{motivation}

This section presents several examples of Euclidean representations of Riemannian manifolds whose Riemannian distances are not Euclidean.  We begin with map projection, the ancient problem of representing the globe (or some subset thereof) in $\Re^2$.  We progress to the representation of closed curves in $\Re^2$, which we then extend to rectangular annuli.  The latter resemble an example in \cite{hlle:2003} that has been widely interpreted as illustrating the limitations of Isomap.

\subsection{Map Projection}
\label{maps}

In mathematical cartography, a {\em map projection}\/ is a function that maps the sphere $S^2 \subset \Re^3$ (or a subset thereof) to $\Re^2$.  It has long been appreciated that map projection necessarily induces distortion.  Hundreds of map projections have been proposed, most with the goal of preserving some salient geometric property.
Equidistant projections preserve some---not all---great circle distances.  (As we shall see in Section~\ref{recovery}, it is impossible to preserve all great circle distances.\footnote{According to \cite[pp.~73--74]{maps:1993}, the ``first formal proof that the surface of a sphere cannot be transformed to a plane without distortion of some sort'' was given by Euler in 1777.})  Conformal projections preserve angles between intersecting great circles.  Authalic projections preserve surface area.  Compromise projections attempt to balance these properties, against each other and/or against other desiderata.  See \cite{maps:1993}, from which most of the material in this section is derived, for a comprehensive survey of the history of map projection.

We begin with three well-known map projections, then consider multidimensional scaling as a compromise.  While it is impossible to find $z_1,\ldots,z_n \in \Re^2$ with Euclidean interpoint distances equal to the great circle interpoint distances of $x_1,\ldots,x_n \in S^2$, minimizing the raw stress criterion comes as close as possible to achieving that goal.  Each of the maps that follow is a Euclidean representation of $n=10000$ points in the Western Hemisphere.  For simplicity, we model the Earth as a sphere; in practice, ellipsoidal models of the Earth are slightly more accurate.  Following \cite{maps:1993}, we denote latitude in radians by $\phi$ and longitude in radians by $\lambda$.  For the formulae that follow, the Western Hemisphere is $(\phi,\lambda) \in [-\pi/2,\pi/2] \times [-\pi,0]$.

\begin{figure}[tb]
\begin{center}
{\includegraphics[width=0.8\textwidth]{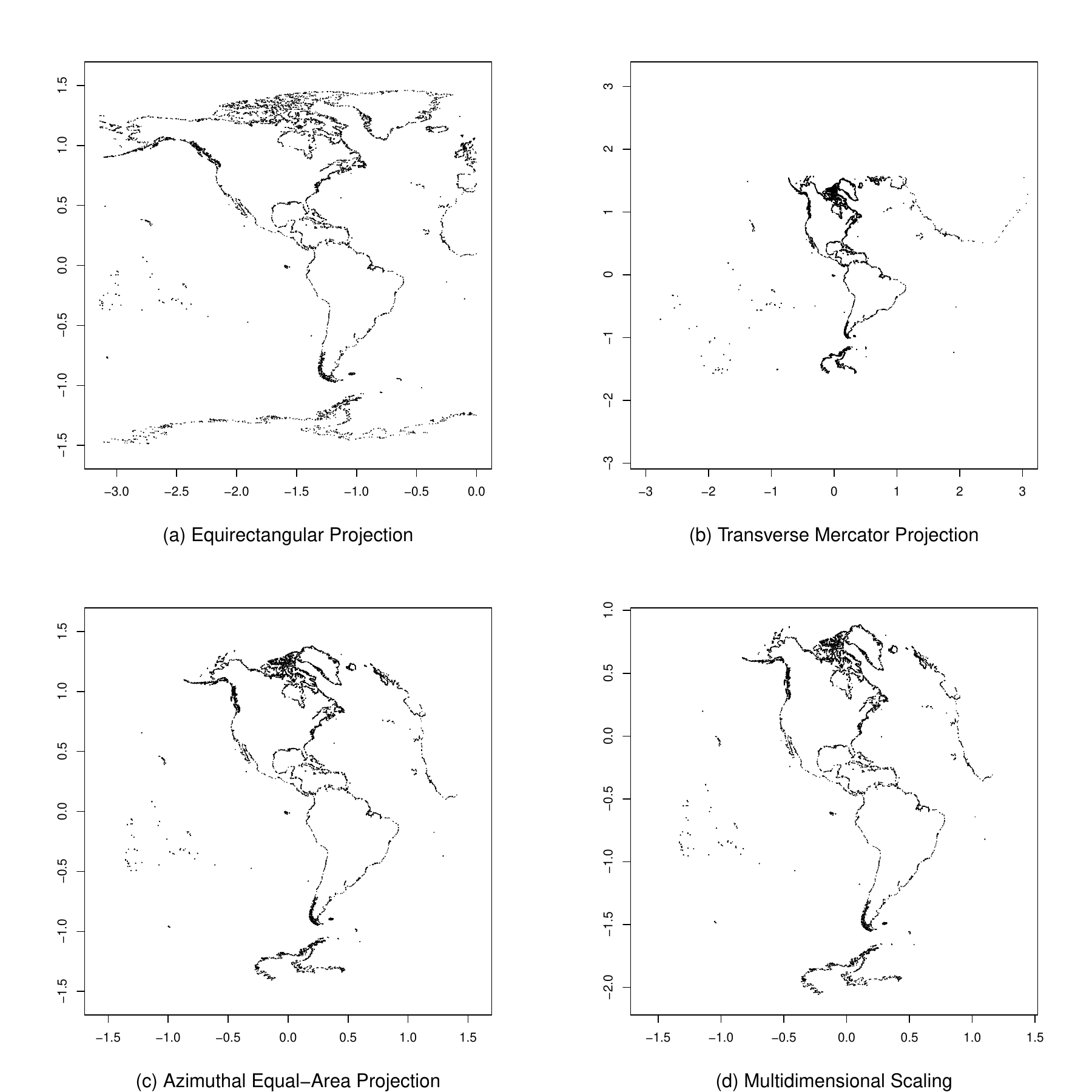}}
\end{center}
\caption{Four map projections of $n=10000$ points on coastlines in the Western Hemisphere.}
\label{fig:WH}
\end{figure}

\begin{example}[Equidistant Projection]
Treating longitude and latitude as Cartesian coordinates,
we obtain $z = R(\lambda,\phi)$, where $R$ determines the scale of the map.  (For our purposes, scale is not important and we henceforth set $R=1$.)  
This is the ancient {\em equirectangular projection}, also called {\em plate carr\'{e}e}\/ and {\em plane chart}.  According to \cite[p.~6]{maps:1993}, ``Ptolemy credited Marinus of Tyre with the invention about A.D.~100.''  It remained popular through the Renaissance, but more sophisticated map projections were discovered in the 1700s, rendering equirectangular projection nearly obsolete by 1800.
Equirectangular projections preserve distance along the equator and along any longitudinal meridian, but severely distort other distances near the poles.  An equirectangular projection of the Western Hemisphere is displayed in Figure~\ref{fig:WH}(a).
\end{example}

\begin{example}[Conformal Projection]
In a remarkable paper published in 1772, J.~H. Lambert proposed seven new map projections of various types.  His third projection, subsequently known as the {\em transverse Mercator projection}, would become the ``leading projection in the 20th Century for large-scale maps.''~\cite[p.~85]{maps:1993}  Defined by
\[
z = R \left( \frac{1}{2} \log \left( \frac{1+ \cos\phi \, \sin \left( \lambda-\lambda_0 \right)}{1- \cos\phi \, \sin \left( \lambda-\lambda_0 \right)} \right), \arctan \left( \frac{\tan\phi}{\cos \left( \lambda-\lambda_0 \right)} \right) \right), 
\]
where $\lambda_0$ specifies which longitudinal meridian will be central, transverse Mercator projections are conformal and represent both the equator and the specified central meridian as straight lines.  A transverse Mercator projection of the Western Hemisphere with $\lambda_0 = -\pi/2$ is displayed in Figure~\ref{fig:WH}(b).
\end{example}

\begin{example}[Authalic Projection]
Lambert also proposed three authalic projections, one of which ``is now commonly seen in atlases.''~\cite[p.~87]{maps:1993}
Technically,
{\em Lambert's azimuthal equal-area projection}\/ is a family of projections indexed by $\phi_0$, the latitude of the center of projection.  Setting $\phi_0=0$, so that the center of projection lies on the equator,
\[
z = R \left[ \frac{2}{1+ \cos\phi \, \cos \left( \lambda-\lambda_0 \right)} \right]^{1/2} \left( \cos\phi \, \sin \left( \lambda-\lambda_0 \right),\sin\phi \right).
\]
An azimuthal equal-area projection of the Western Hemisphere with $\lambda_0 = -\pi/2$ is displayed in Figure~\ref{fig:WH}(c).
\end{example}

\begin{example}[Multidimensional Scaling]
Suppose that $x_i = (\phi_i,\lambda_i)$ and $x_j = (\phi_j,\lambda_j)$ lie on the unit sphere.  Let $y_i$ and $y_j$ represent $x_i$ and $x_j$ by Cartesian coordinates in $\Re^3$, via the transformation
\[
y = \left( \sin (\phi+\pi/2) \, \cos \lambda, \;
\sin (\phi+\pi/2) \, \sin \lambda, \;
\cos (\phi+\pi/2) \right).
\]
The great circle distance between $x_i$ and $x_j$ is $\delta_{ij} = \arccos \, \langle y_i,y_j \rangle$.
Given $n$ such points, let $\Delta=[\delta_{ij}]$ and find $z_1,\ldots,z_n \in \Re^2$ that minimize (\ref{eq:stress}).
The resulting map will be neither conformal nor authalic, nor will it be exactly equidistant along longitudinal meridians; however it will be as nearly equidistant as possible in the sense of a plausible error criterion.  (Notice that, by carefully choosing the $w_{ij}$ in (\ref{eq:stress}), one can control which regions of the map are more or less distorted.)  Because the $\delta_{ij}$ contain no sense of compass direction, the $z_i$ may have to be reflected and/or rotated to obtain a conventional orientation.
Such a map of the Western Hemisphere is displayed in Figure~\ref{fig:WH}(d).
\end{example}

We are not proposing multidimensional scaling as an alternative to traditional map projection; however, the following observations are crucial to our development:
\begin{enumerate}

\item  Traditional map projection concedes that a completely faithful representation of a hemisphere in $\Re^2$ is impossible.

\item  Traditional map projection exploits a detailed understanding of spherical geometry.

\item  Multidimensional scaling achieves a plausible Euclidean representation of a hemisphere using only pairwise great-circle distances.

\end{enumerate}
In light of these observations, there is an obvious way to proceed if one is presented with points that lie on an unknown Riemannian manifold: estimate the pairwise Riemannian distances, then use multidimensional scaling to embed the estimated Riemannian distances in $\Re^d$.  This is precisely what Isomap does.  Isomap cleverly estimates Riemannian distance, but the problem of approximating a non-Euclidean distance with Euclidean distance is unavoidable.

\subsection{Closed Curves} 
\label{curves}

\begin{figure}[ht]
\begin{center}
{\includegraphics[width=0.5\textwidth]{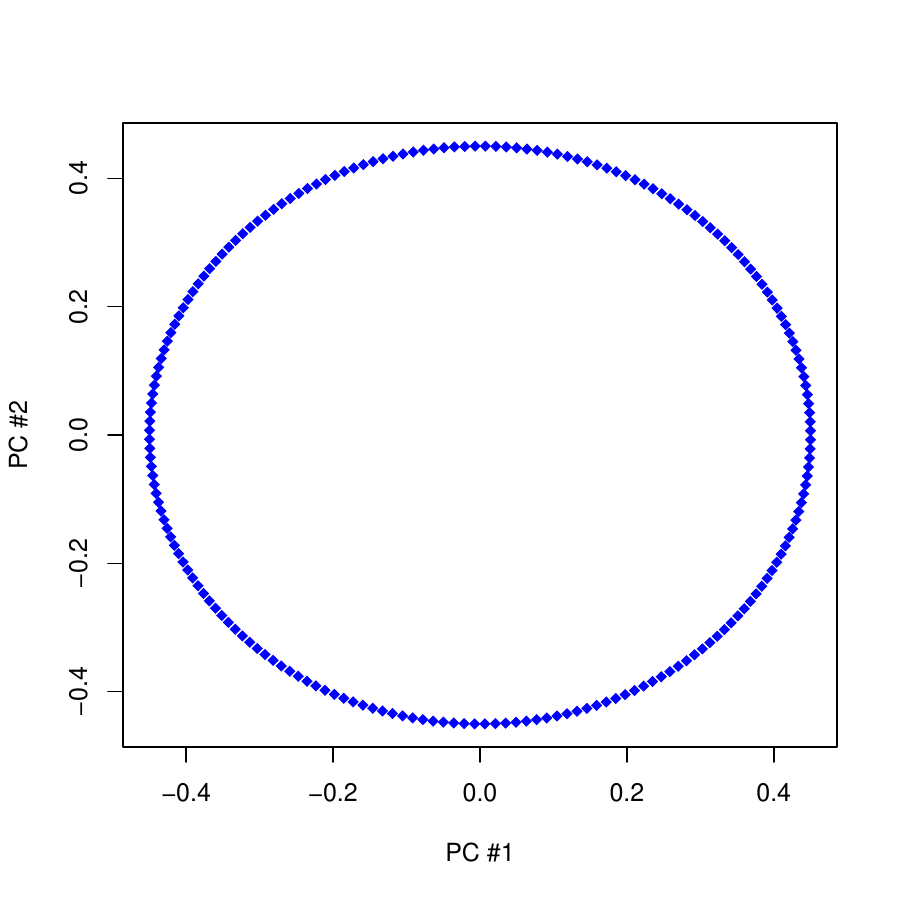}}
\end{center}
\caption{CMDS embedding of $200$ equally spaced points on a closed curve of length $2$.  The matrix of pairwise arc distances, hence the embedding, does not depend on the shape of the curve.}
\label{fig:GRep-motiv}
\end{figure}

The main point of this section is that all closed curves have the same (non-Euclidean) metric structure.  We make our case by comparing two specific closed curves.
\begin{example}[Two Closed Curves]
Consider (1) the rectangle $R \subset \Re^2$ with vertices at $(\pm 0.05,0.05)$ and $(\pm 0.05,0.95)$, and (2) the circle $S \subset \Re^2$, centered at $(0,0)$ with radius $1/\pi$.  The perimeter of $R$ has a total length of $2$; similarly, the circumference of $S$ is $2$.
For $x,y \in R$, let $\delta_R(x,y)$ denote the length of the shortest arc in $R$ that connects $x$ and $y$; for $x,y \in S$, let $\delta_S(x,y)$ denote the length of the shortest arc in $S$ that connects $x$ and $y$.

Let $n=200$.  Let $x_1,\ldots,x_n \in R$ be equally spaced with respect to $\delta_R$ and let $y_1,\ldots,y_n \in S$ be equally spaced with respect to $\delta_S$.  (For example, place $x_1$ at $(0.05,0.05)$ and place $x_2,\ldots,x_n$ counterclockwise at increments of $0.01$.  Place $y_1$ at $(1/\pi,0)$ and place $y_2,\ldots,y_n$ counterclockwise at increments of $0.01$.)  Let $X$ and $Y$ denote the corresponding configuration matrices and define dissimilarity matrices
\begin{eqnarray*}
D(X) = \left[ \left\| x_i-x_j \right\| \right] & \mbox{ and } &
D(Y) = \left[ \left\| y_i-y_j \right\| \right]; \\
\Delta(X) = \left[ \delta_R \left( x_i,x_j \right) \right] & \mbox{ and } &
\Delta(Y) = \left[ \delta_S \left( y_i,y_j \right) \right]. 
\end{eqnarray*}

Let us say that two configurations are isometric if their matrices of interpoint distances are identical.
By definition, both $D(X)$ and $D(Y)$ are EDM-1.  Hence, a configuration that is isometric to $X$ can be recovered from $D(X)$ and a configuration that is isometric to $Y$ can be recovered from $D(Y)$.  For example, CMDS recovers a configuration whose centroid lies at the origin and whose coordinate axes are the principal components of the configuration.  Because $X$ and $Y$ are not isometric, $D(X) \neq D(Y)$ and the recovered configurations are not isometric.

By construction, $\Delta(X)=\Delta(Y)=\Delta$.  The metric structures of $R$ and $S$ defined by $\delta_R$ and $\delta_S$ are identical: it is not possible to recover from $\Delta$ the distinction between the rectangular shape of $R$ and the circular shape of $S$.  Furthermore, applying Theorem~\ref{thm:embed.e}, we discover that $\Delta$ is not EDM-1.  The $n \times n$ matrix $\tau(\Delta_2)$ has $100$ positive eigenvalues, one zero eigenvalue, and $99$ negative eigenvalues.  The negative eigenvalues correspond to the non-Euclidean portion of $\Delta$ and have a total variation of $16.665$.  The positive eigenvalues correspond to the Euclidean portion of $\Delta$ and have a total variation of $50$.  The first two principal components of this $100$-dimensional Euclidean configuration explain $40.53181/50 \doteq 81\%$ of its total variation.  This $2$-dimensional configuration of points is displayed in Figure~\ref{fig:GRep-motiv}.
\label{ex:curves}
\end{example}

One's initial impression of Figure~\ref{fig:GRep-motiv} is that CMDS has recovered $Y$ but not $X$.  This impression is misleading, because Figure~\ref{fig:GRep-motiv} was constructed solely from $\Delta$.  The pairwise arc distances are the same for $Y$ and $X$; hence, the representation in Figure~\ref{fig:GRep-motiv} is equally valid for $Y$ and for $X$.  The key to understanding Figure~\ref{fig:GRep-motiv} is appreciating that it is a Euclidean approximation of a non-Euclidean structure.

First, the circle on which the points in Figure~\ref{fig:GRep-motiv} lie is {\em not}\/ $S$.  Its radius is approximately $0.45$, not $1/\pi \doteq 0.32$.  Second, the $2$-dimensional configuration in Figure~\ref{fig:GRep-motiv} is only an approximation, the projection of a $100$-dimensional configuration onto its first two principal components.  Third, even the $100$-dimensional configuration is only an approximation, specifically the best least squares approximation of $\tau(\Delta_2)$ by centered Euclidean inner products.

Properly interpreted, it makes perfect sense that equally spaced points on any closed curve would lead CMDS to construct a circular configuration of points from the pairwise arc distances.  If the matrix of equally spaced arc distances is $\Delta$, then the corresponding matrix of (fallible) centered inner products $B = \tau(\Delta_2)$ has constant diagonal entries of $b^2$, suggesting that all points should be placed on a sphere of radius $b$.  Moreover, the matrix of (fallible) angles is $A = [ \arccos ( b_{ij}/b^2 )]$, and the angles between each pair of consecutive points have a constant value $a$.  Thus viewed, a circular configuration of points is the obvious $2$-dimensional embedding of $\Delta$.

Although embedding $\Delta$ in $\Re^2$ (or in any $\Re^d$) does not {\em recover}\/ $X$ or $Y$, the representation of $\Delta$ in Figure~\ref{fig:GRep-motiv} is of evident value.
However, it must be emphasized that Figure~\ref{fig:GRep-motiv} represents the metric structure of $R$ and $S$ by $2$-dimensional {\em Euclidean}\/ structure.  The arc distances measured by $\delta_R$ and $\delta_S$ are approximated by Euclidean distances in Figure~\ref{fig:GRep-motiv}.  Although the points in Figure~\ref{fig:GRep-motiv} lie on a circle, it is the chordal distances between these points that approximate the arc distances in $\Delta$.

\subsection{Rectangular Annuli}
\label{annuli}
Example~\ref{ex:curves} described two closed curves, $R$ and $S$, in $\Re^2$.  We now focus on $R$, but we introduce borders to create  $2$-dimensional manifolds whose metric structure resembles the metric structure of $R$.

\begin{figure}[ht]
\begin{center}
{\includegraphics[width=1\textwidth]{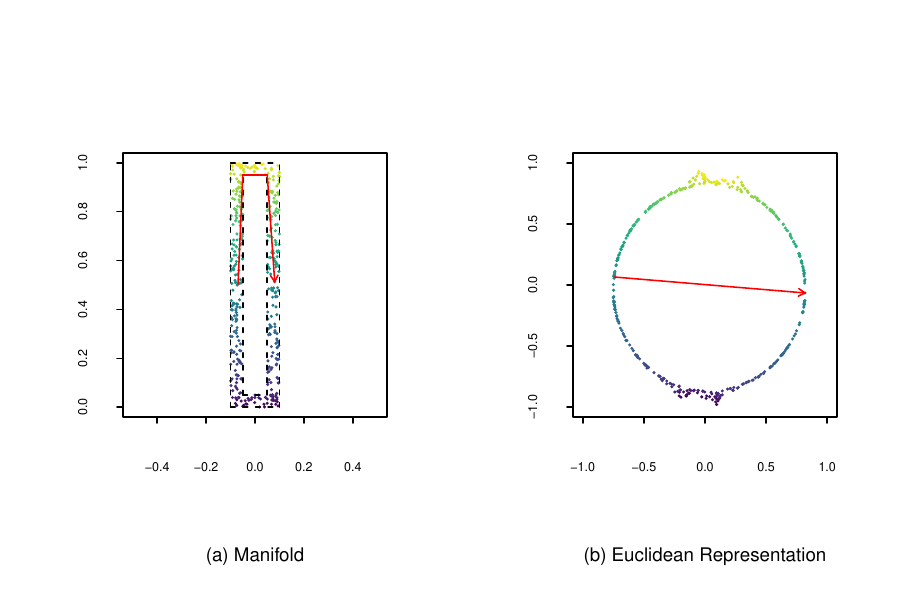}}
\end{center}
\caption{Example~\ref{ex:RA1}.  (a) $n=400$ points on a rectangular annulus indicated by dashed black lines.  The red arrow indicates a geodesic curve on the manifold. (b) Euclidean representation of their pairwise Riemannian distances, obtained by minimizing the raw stress criterion.  The red arrow indicates the corresponding geodesic curve in $\Re^2$.}
\label{fig:RA1}
\end{figure}

\begin{example}
Define rectangles $R_1 \subset \Re^2$ with vertices $\{(\pm 0.05, 0.05),(\pm 0.05,0.95)\}$, and
$R_2 \subset \Re^2$ with vertices $\{(\pm 0.1,0),(\pm 0.1,1)\}$.  Let $\M$ denote the subset of $\Re^2$ that lies outside $R_1$ but inside $R_2$.  In analogy with a traditional annulus, we describe $\M$ as a {\em rectangular annulus}.  Like $R_1$, the metric structure of $\M$ is non-Euclidean.  Considering how closely $\M$ resembles $R_1$, we would expect Euclidean representations of their respective metric structures to closely resemble each other.

Let $n=400$.  We randomly generated $x_1,\ldots,x_n \iid \mbox{\rm Uniform}(\M)$, displayed in Figure~\ref{fig:RA1}(a), and computed the pairwise Riemannian distances $\Delta=[\delta_{ij}]$ between them.  We then embedded $\Delta$ in $\Re^2$ by minimizing (\ref{eq:stress}) with $w_{ij}=1$, following CMDS with $20$ iterations of the Guttman transformation.  This resulted in the configuration displayed in Figure~\ref{fig:RA1}(b).  Note the expected strong resemblance between Figures \ref{fig:RA1}(b) and \ref{fig:GRep-motiv}.
\label{ex:RA1}
\end{example}

The rectangular annulus in Example~\ref{ex:RA1} closely resembles $R$ in Example~\ref{ex:curves}.  Modifying $R_1$ and $R_2$, we obtain another rectangular annulus that more closely resembles an example in \cite{hlle:2003}.

\begin{figure}[ht]
\begin{center}
{\includegraphics[width=1\textwidth]{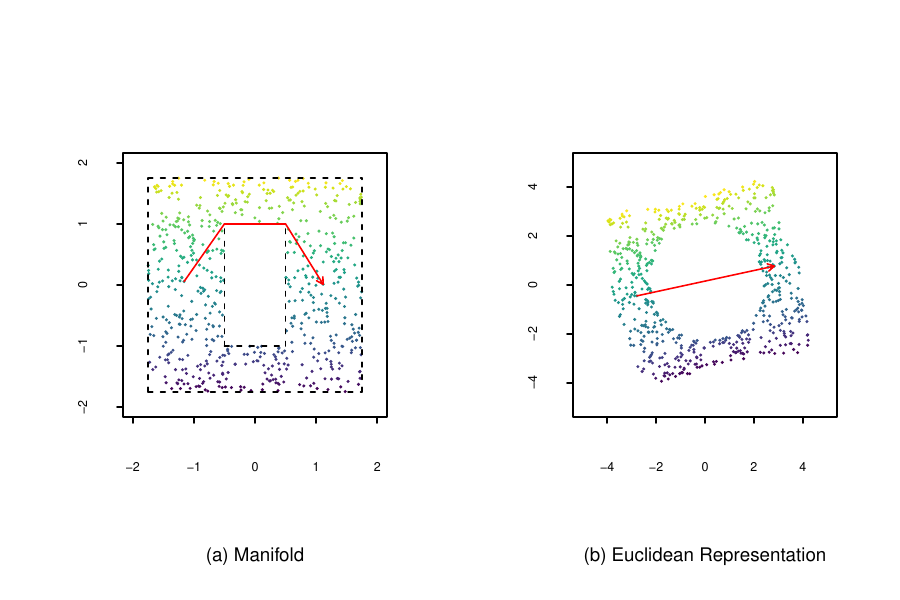}}
\end{center}
\caption{Example~\ref{ex:RA2}.  (a) $n=600$ points on a rectangular annulus indicated by dashed black lines.  The red arrow indicates a geodesic curve on the manifold.  (b) Euclidean representation of their pairwise Riemannian distances, obtained by minimizing the raw stress criterion.  The red arrow indicates the corresponding geodesic curve in $\Re^2$.}
\label{fig:RA2}
\end{figure}

\begin{example}
Define rectangles $R_1 \subset \Re^2$ with vertices 
\[
\{(-0.25,-1.2),  (0.55,-1.2), (-0.25,0.3), (0.55,0.3)\},
\] 
and $R_2 \subset \Re^2$ with vertices $\{(\pm 1.75,-2), (\pm 1.75,1.5)\}$.  Let $\M$ denote the subset of $\Re^2$ that lies outside $R_1$ but inside $R_2$.  Let $n=600$.  We randomly generated $x_1,\ldots,x_n \iid \mbox{\rm Uniform}(\M)$, displayed in Figure~\ref{fig:RA2}(a), and computed the pairwise Riemannian distances $\Delta=[\delta_{ij}]$ between them.  We then embedded $\Delta$ in $\Re^2$ by minimizing (\ref{eq:stress}) with $w_{ij}=1$, following CMDS with $20$ iterations of the Guttman transformation.  This resulted in the configuration displayed in Figure~\ref{fig:RA2}(b).
Note the resemblance between Figures \ref{fig:RA2}(b), \ref{fig:RA1}(b), and \ref{fig:GRep-motiv}.
\label{ex:RA2}
\end{example}

Figures \ref{fig:GRep-motiv}, \ref{fig:RA1}(b), and \ref{fig:RA2}(b)  resemble each other because each of the manifolds in Examples \ref{ex:curves}, \ref{ex:RA1} and \ref{ex:RA2} contain geodesic curves that bend around a central region of ambient space.   The corresponding Euclidean representations of metric structure attempt to approximate these curves as straight lines, resulting in what appears to be a warping of the original structure.

\section{Isomap}
\label{isomap}

\begin{figure}[tb]
\begin{center}
\fbox{ 
\begin{minipage}{5in}
\vspace{1em}
Given: feature vectors $x_1,\ldots,x_n \in {\mathcal M} \subset \Re^q$ and a target dimension $d$.
\begin{enumerate}

\item Construct an $\epsilon$-neighborhood or $K$-nearest-neighbor graph of the observed feature vectors.
Weight edge $j \leftrightarrow k$ of the graph by $\| x_j-x_k \|$.

\item Compute the dissimilarity matrix $\Delta = [ \delta_{jk} ]$,
where $\delta_{jk}$ is the shortest path distance between vertices $j$ and $k$.  The key idea that underlies Isomap is that shortest
path distances on a locally connected graph approximate 
Riemannian distances on an underlying Riemannian manifold ${\mathcal M}$.

\item Embed $\Delta$ in $\Re^d$.  
Traditionally, Isomap embeds by classical multidimensional scaling (CMDS);
however, if one's goal is to approximate shortest path distance with
Euclidean distance, then one might prefer to embed differently, e.g., by minimizing Kruskal's \cite{kruskal:1964a} raw stress criterion.

\end{enumerate}
\vspace{0.5em}
\end{minipage}
}
\end{center}
\caption{Isomap, the manifold learning procedure proposed in \cite{isomap:2000}.}
\label{fig:isomap}
\end{figure}

The Isomap procedure for manifold learning is summarized in 
Figure~\ref{fig:isomap}.
The key idea that underlies Isomap is that shortest path distances on a locally connected graph approximate 
Riemannian distances on an underlying Riemannian manifold.   For that reason, it seems natural to embed by minimizing an error criterion such as the raw stress criterion that measures how well the embedded Euclidean interpoint distances approximate the shortest path distances.  In fact, the authors of \cite{isomap:2000} elected to embed the shortest path distances by CMDS.  Because the originality of Isomap lies in its approximation of Riemannian distances with shortest path distances, our view is that any embedding of shortest path distances is appropriately described as Isomap.\footnote{Some have described Isomap as an {\em extension}\/ of CMDS, but the embedding step in Isomap is routine.  We regard Isomap as an ingenious {\em application}\/ of CMDS---or of some other embedding technique.}  Notice that
Isomap poses two model selection problems for the user: the choice of localization parameter ($\epsilon$ or $K$), and the choice of target dimension ($d$).

In an intriguing paragraph in \cite{isomap:2000}, the authors attempt to identify and reconcile two conceptually distinct interpretations of Isomap:
\begin{quote}
``Just as PCA and MDS are guaranteed, given sufficient data, to recover the true structure of linear manifolds, Isomap is guaranteed asymptotically to recover the true 
dimensionality and geometric structure of a strictly larger class of nonlinear manifolds. Like the Swiss roll, these are manifolds
whose intrinsic geometry is that of a convex region of Euclidean space, but whose 
ambient geometry in the high-dimensional input space may be highly folded, twisted, or curved. For non-Euclidean manifolds, such as a hemisphere or the surface of a doughnut, Isomap still produces a globally optimal low-dimensional Euclidean representation, as measured by Eq.\ 1.''
\end{quote}
The authors do not specify precisely what it means to recover geometric structure, nor the class of nonlinear manifolds for which recovery by Isomap is guaranteed, but ``these are manifolds
whose intrinsic geometry is that of a convex region of Euclidean space.''  This interpretation of manifold learning was formally stated in \cite{hlle:2003} as the Parametrization Recovery Problem, which we discuss in Section~\ref{recovery}.

The authors' reference to convexity is intriguing.  Convexity is indeed required for Isomap to solve the Parametrization Recovery Problem, but we demonstrate in Section~\ref{converge} that convexity is not needed to ensure that shortest path distances converge to Riemannian distances.  Thus, whether or not Isomap requires convexity depends on whether one is trying to recover parameters or trying to represent geodesic structure in Euclidean space.

For ``non-Euclidean manifolds,'' the authors contend that Isomap still produces something reasonable, in the sense of Theorem~\ref{thm:psd}.  We agree, although we would argue that it would be more reasonable still to minimize the raw stress criterion and directly approximate shortest path distances with Euclidean distances.  Whatever the embedding technique, what Isomap is producing is a Euclidean representation of geodesic structure.  In our view, producing such should not be regarded as a consolation prize when Isomap fails to recover parameters, but as an equally legitimate end in itself.

How one interprets the phrase ``non-Euclidean manifold'' is critical.  If ``Euclidean'' means that $\M$ is locally isometric to Euclidean space, then parametrization recovery is possible, as demonstrated in \cite{hlle:2003} and discussed in Section~\ref{recovery}.  If ``Euclidean'' means that $\M$ is globally isometric to Euclidean space, then parametrization recovery is possible and represents the geodesic structure of $\M$ exactly. 
However, if $\M$ is locally but not globally isometric to Euclidean space (as with the rectangular annuli of Section~\ref{annuli}), then the objectives of parametrization recovery and geodesic representation are in tension.  For such manifolds, it may be technically correct to say that Isomap fails at parametrization recovery, but we find it more instructive to say that Isomap succeeds (approximately) at geodesic representation.

\section{Parametrization Recovery}
\label{recovery}


The concept of parametrization recovery appears in \cite{isomap:2000},
in the claim that ``Just as PCA and MDS are guaranteed, given sufficient data, to recover the true structure of linear manifolds, Isomap is guaranteed asymptotically to recover the true dimensionality and geometric structure of a strictly larger class of nonlinear manifolds.''  The Parametrization Recovery Problem was informally stated by 
Donoho and Grimes \cite{hlle:2003} as follows:
\begin{quote}
Let $\Theta \subset \Re^p$ be a parameter space and let
$\psi: \Theta \rightarrow \Re^q$ be a smooth injection.
Given feature vectors $m_i = \psi(\theta_i) \in \M = \psi(\Theta)$,
recover the mapping $\psi$ and the parameter points $\theta_i$.
\end{quote}
They noted that this statement of the problem is ill-posed,
requiring additional assumptions in order to uniquely determine solutions.

Isomap can (in theory) recover manifolds that are globally isometric to a convex subset of Euclidean space.  Note that such manifolds are necessarily connected.  Donoho and Grimes relaxed this condition:
their technique of Hessian eigenmaps (``Hessian LLE'') can (in theory) recover connected manifolds that are locally (not necessarily globally) isometric to Euclidean space.
They introduced a quadratic form, ${\mathcal H}$, defined on $C^2$ (continuously twice differentiable) functionals on a manifold $\M=\psi(\Theta)$,
where $\Theta$ is an open connected subset of $\Re^p$ and
$\psi$ is a locally isometric embedding of $\Theta$ into $\Re^q$.
Their key result states that ``the original isometric coordinates $\theta$
can be recovered, up to a rigid motion, by identifying a suitable basis for the
null space of ${\mathcal H}$.''  This is parametrization recovery.  Hessian eigenmaps are constructed from discrete approximations of ${\mathcal H}$.

Is it constructive to interpret manifold learning as parametrization recovery?  Following \cite{Lee:1997},
a Riemannian manifold is {\em flat}\/ if and only if it is locally isometric to an open subset of Euclidean space.  All is well if $d=1$, for {\em every}\/ $1$-dimensional Riemannian manifold is 
flat \cite[p.\ 116]{Lee:1997}.  
Thus, spirals can be straightened and their parametrizations recovered.

The case of $d=2$ and $q=3$ is the subject of classical differential geometry.
A $2$-dimensional Riemannian manifold embedded in $\Re^3$ is a {\em surface}.
The {\em Gaussian curvature}\/ of a surface ${\mathcal M}$ at $m$ is the product of the
principal curvatures at $m$: $K(m) = \kappa_1(m) \kappa_2(m)$.
If $K(m)=0$, then there is an arc in ${\mathcal M}$ through $m$ that 
is a straight line in $\Re^3$.  A Swiss roll has constant zero curvature (it curves in one principal direction and not in the other), whereas
a hemisphere has constant positive curvature (it curves in both principal directions).

Gauss's celebrated {\em Theorema Egregium}\/ (1827) states that Gaussian curvature is invariant under local isometry.  
Hence, if ${\mathcal M}$ is locally isometric to some
$\Theta \subset \Re^2$, then ${\mathcal M}$ must have constant zero Gaussian curvature.  Thus, the {\em only}\/ surfaces for which parametrization recovery is possible are those that curve in at most one principal direction. 
Parametrization recovery is possible for Swiss rolls, but not for hemispheres. 

In the general case, ``A Riemannian manifold is flat if and only if its curvature tensor vanishes identically.'' \cite[Theorem 7.3]{Lee:1997}
The curvature tensor at $m \in {\mathcal M}$ is completely determined by the sectional curvatures at $m$,
i.e., the Gaussian curvatures at $m$ of the $2$-dimensional submanifolds at $m$ that are swept out by geodesics whose initial tangent vectors lie in a $2$-dimensional subspace of the tangent space of ${\mathcal M}$ at $m$.  See \cite[Chapter 8]{Lee:1997} for details.  It follows that ${\mathcal M}$ is flat if and only if each sectional curvature at every $m \in {\mathcal M}$ is zero.  But this means that, at any point in a flat manifold, there can be at most one principal direction in which the manifold curves.  Thus, there are no manifolds with curvature more complicated than a Swiss roll for which parametrization recovery is possible.  Small wonder that Swiss rolls appear so frequently in the manifold learning literature!

An example in \cite[Section 7]{hlle:2003} considers data sampled from a modified Swiss roll:
\begin{quote}
``Instead of sampling parameters in a full rectangle, we sample from a rectangle with a missing rectangular strip punched out of
the center.  The resulting Swiss roll is then missing the 
corresponding strip and thus is not convex (while still remaining connected).''
\end{quote}
This is an example of a rectangular annulus, described in Section~\ref{annuli}.  Referring to their Figure~1, Donoho and Grimes observed that
\begin{quote}
``In the case of ISOMAP, the nonconvexity causes a strong dilation of the missing region, warping the rest of the embedding. Hessian LLE, on the other hand, embeds the result almost perfectly into two-dimensional space.''
\end{quote}
From the perspective of parametrization recovery, this interpretation is completely correct.  From the perspective of geodesic representation, however, what Donoho and Grimes regarded as a ``missing region [that] warp[s] the rest of the embedding'' is in fact the appropriate Euclidean representation of geodesic structure, illustrated by the examples in Section~\ref{annuli}.  That Isomap constructed this representation from shortest path distances rather than actual Riemannian distances might equally well be interpreted as a resounding success.

\section{Convergence Analysis}
\label{converge}

We now consider under what circumstances shortest path distances approximate Riemannian distances.
This section follows the analysis in \cite{Bernstein&etal:2000}, clarifying several ambiguities.  Of particular interest, the authors write that ``We say that $\M$ is {\em geodesically convex}\/ if any two points $x,y$ in $\M$ are connected by a geodesic of length $d_\M(x,y)$.''  Every compact connected Riemannian manifold has this property, so assuming it is unnecessary.  Unfortunately, their terminology reinforces the impression that Isomap requires convexity, and their analysis omits details that might have mitigated misinterpretation.
The following analysis makes explicit the reasoning in \cite{Bernstein&etal:2000}, demonstrating that convexity is not required for convergence.

Let $\M \subset \Re^q$ be a compact connected $d$-dimensional Riemannian manifold.  Following \cite{Bernstein&etal:2000},
the {\em minimum radius of curvature}\/ of $\M$, $r_0$, is defined by
\[
1/r_0 = \max_{\gamma, t} \left\| \ddot{\gamma}(t) \right\|,
\]
where $\gamma$ varies over all unit-speed geodesic curves in $\M$ and $t$ varies over the domain of $\gamma$. 
The {\em minimum branch separation}\/ of $\M$, $s_0$, is the largest positive number for which $\| x-y \| <s_0$ entails $d_\M(x,y) \leq \pi r_0$ for any $x,y \in \M$.
Both $r_0$ and $s_0$ necessarily exist because $\M$ is compact.

The following inequality appears in \cite[Appendix]{Bernstein&etal:2000} as the Minimum Length Lemma, accompanied by a remark that ``We expect that there is a shorter proof of the Minimum Length Lemma using calculus.''  Such a proof was suggested to us by Bruce Solomon.

\begin{lemma}[Minimum Length]
Let $\gamma \colon [-\ell/2,\ell/2] \to \Re^q$ be a smooth arc in $\M$. 
Suppose that $\gamma$ is parametrized by arc length, i.e., 
$\| \dot{\gamma}(t) \| = 1$, 
and satisfies
$\| \ddot{\gamma}(t) \| \leq 1/r_0$ 
for every $t \in [-\ell/2,\ell/2]$. If $\mbox{\rm length}(\gamma) = \ell \leq \pi r_0$, then
\[ 
\left\| \gamma (\ell/2) - \gamma (-\ell/2) \right\| \geq 2r_0 \sin \left( \ell/2r_0 \right)  .
\]
\label{lm:MinLength}
\end{lemma}
\subparagraph{Proof}
Because each $\| \dot{\gamma}(t) \| = 1$, 
$\dot{\gamma}$ maps $[-\ell/2,\ell/2]$ onto the unit sphere $S^{q-1}$.
On $S^{q-1}$, for any $-\ell/2 \leq t_1 \leq t_2 \leq \ell/2$, 
the ``great circle'' distance between $\dot{\gamma}(t_1)$ and $\dot{\gamma}(t_2)$ satisfies
\[
d_{S^{q-1}} \left( \dot{\gamma} \left( t_1 \right),\dot{\gamma} \left( t_2 \right) \right) \leq \mbox{\rm length} \left( \dot{\gamma} \left( \left[ t_1,t_2 \right] \right) \right) = \int_{t_1}^{t_2} \left\| \ddot{\gamma}(t) \right\| \, dt \leq \left| t_2-t_1 \right| /r_0. 
\]
For $t \in [-\ell/2,\ell/2]$, let
$\psi(t)= d_{S^{n-1}}(\dot{\gamma}(0),\dot{\gamma}(t))$.  Because $\psi(t)$ is also the angle between $
\dot{\gamma}(0)$ and $\dot{\gamma}(t)$, we have 
$\cos \psi(t) = \langle \dot{\gamma}(0), \dot{\gamma}(t) \rangle$.

Notice that $0 \leq \psi(t) \leq  |t-0|/r_0 = |t|/r_0 \leq \pi/2$.
Recall that $\cos u$ is increasing on $[-\pi/2,0]$ and decreasing in $[0,\pi/2]$. Because $\psi(t) \geq 0 \geq t/r_0$ for $t \in [-\ell/2,0]$ and $\psi(t) \leq t/r_0$ for $t \in [0,\ell/2]$, we obtain
$\cos \psi(t) \geq \cos \left( t/r_0 \right)$ for all $t \in [-\ell/2,\ell/2]$. 
Applying the Cauchy-Schwartz Inequality,
\begin{eqnarray*}
    \left\| \gamma(\ell/2) - \gamma(-\ell/2) \right\| & \geq & \left| \left\langle \gamma(\ell/2) - \gamma(-\ell/2),  \dot{\gamma}(0) \right\rangle \right|  = \int_{-\ell/2}^{\ell/2} \left\langle \dot{\gamma}(t), \dot{\gamma}(0) \right\rangle \ dt \\ 
    & = & \int_{-\ell/2}^{\ell/2}\cos{\psi(t)}\ dt 
    \geq \int_{-\ell/2}^{\ell/2}\cos\left( t/r_0 \right)\ dt = 2r_0 \sin \left( \ell/2r_0 \right).
\end{eqnarray*}
\hfill $\Box$

\medskip

Given $\delta>0$, suppose that the finite set $V \subset \M$ satisfies the {\em $\delta$-sampling condition}, i.e., for each $x\in \M$ there exists $x_i \in V$ for which $d_\M(x,x_i)<\delta$. Given $\epsilon>0$, let
$\G=(V,E)$ be a graph with vertex set $V$ and edges between $x_i$ and $x_j$ if and only if $\| x_i-x_j \| \leq \epsilon$.  Assuming that $\epsilon$ has been chosen so that $\G$ is connected, Bernstein et al.\ \cite{Bernstein&etal:2000} defined two metrics on $\G$:
\begin{eqnarray*}
    d_G(x,y) = \min_P \sum_{i=1}^k \left\| x_{i-1}-x_i \right\|  & \mbox{ and } &
    d_S(x,y) = \min_P \sum_{i=1}^k d_\M \left( x_{i-1},x_i \right),
\end{eqnarray*}
where $P=(x_0,\ldots,x_k)$ varies over all paths along the edges of $\G$ with $x=x_0$ and $y=x_k$.  Because $\|x-y\| \leq d_\M(x,y)$, $d_G(x,y) \leq d_S(x,y)$ for all $x,y \in V$.

The following result is analogous to Main Theorem~A in \cite{Bernstein&etal:2000}.  For simplicity, we set $\epsilon_{\min} = \epsilon_{\max} = \epsilon$ and $\lambda_1=\lambda_2=\lambda$.
In contrast to Main Theorem~A, we do not assume geodesic convexity.
\begin{theorem}
Let $\M$, $V$, and $\G$ be as described above, with 
\begin{eqnarray*}
\epsilon<s_0, & \epsilon \leq (2/\pi) r_0 \sqrt{24\lambda}, & \mbox{and } \delta \leq \lambda \epsilon /4
\end{eqnarray*} 
for some $\lambda \in (0,1)$. Then 
\begin{equation}
(1-\lambda)d_\M(x,y)\leq d_G(x,y)\leq (1+\lambda)d_\M(x,y)
\label{eq:A}
\end{equation}
for every $x,y \in V \subset \M$.
\label{thm:A}
\end{theorem}
\subparagraph{Proof}
To establish the upper bound in (\ref{eq:A}),
first suppose that
$\ell = d_\M(x,y) < \epsilon$.  Then $x,y \in V$ are connected by an edge in $\G$ and $d_G(x,y)=\|x-y\| \leq \ell < (1+\lambda)\ell$.

If $\ell \geq \epsilon$, then let $\gamma$ denote the minimizing geodesic curve between $x$ and $y$.  Partition $\gamma$ into $k$ segments $(\gamma_{i-1},\gamma_i)$ of lengths
\[
\ell_1 = d_\M \left( \gamma_{i-1},\gamma_i \right) = \epsilon-2\delta
\]
for $i=2,\ldots,k-1$,
and
\[
(\epsilon-2\delta)/2 \leq
\ell_0 = d_\M \left( \gamma_0=x,\gamma_1 \right) = 
d_\M \left( \gamma_{k-1},\gamma_k=y \right) \leq \epsilon-2\delta.
\]
For each $\gamma_2,\ldots,\gamma_{k-1} \in \M$, choose $x_i \in V$ such that $\| x_i - \gamma_i \| < \delta$.  Then
\begin{eqnarray}
d_\M \left( x_{i-1},x_{i} \right) & \leq & 
d_\M \left( x_{i-1},\gamma_{i-1} \right) + 
d_\M \left( \gamma_{i-1},\gamma_{i} \right) +
d_\M \left( \gamma_{i},x_{i} \right) \nonumber \\  & \leq &
 \delta + (\epsilon - 2\delta) +\delta = \epsilon 
  = \ell_1\frac{\epsilon}{\epsilon - 2\delta}
\label{eq:ell1}
\end{eqnarray}
for $i=2,\ldots,k-1$.  Also,
\begin{eqnarray}
d_\M \left( x,x_1 \right) & \leq & 
d_\M \left( x=\gamma_0,\gamma_1 \right) +
d_\M \left( \gamma_1,x_1 \right) \leq 
\ell_0 +\delta \nonumber \\ & = &
\ell_0 \left( 1 + \frac{\delta}{\ell_0} \right) \leq
\ell_0 \left( 1 + \frac{2\delta}{\epsilon-2\delta} \right) =
\ell_0\frac{\epsilon}{\epsilon - 2\delta},
\label{eq:ell0x}
\end{eqnarray}
and likewise
\begin{equation}
d_\M \left( x_{k-1},y \right) \leq
\ell_0\frac{\epsilon}{\epsilon - 2\delta}.
\label{eq:ell0y}
\end{equation}
Finally, note that $\lambda \in (0,1)$ entails
\[
(1-\lambda/2)(1+\lambda) = 1+\lambda-\lambda/2-\lambda^2/2 =
1+(\lambda-\lambda^2)/2 > 1,
\]
and therefore
\begin{equation}
\frac{1}{1-\lambda/2} < 1+\lambda.
\label{eq:lambda}
\end{equation}
Combining (\ref{eq:ell0x}), (\ref{eq:ell1}), (\ref{eq:ell0y}), 
and (\ref{eq:lambda}), we obtain
\begin{eqnarray*}
d_G(x,y) & \leq & d_S(x,y) \leq \sum_{i=1}^{k} d_\M \left( x_{i-1},x_{i} \right) \leq
\frac{\epsilon}{\epsilon - 2\delta} \left[
\ell_0 + (k-2)\ell_1 + \ell_0 \right] \\ & = &
\frac{\epsilon}{\epsilon - 2\delta} \ell \leq
\frac{\epsilon}{\epsilon-\lambda(\epsilon/2)} \ell =
\frac{1}{1-\lambda/2} \ell <
(1+\lambda) \ell.
\end{eqnarray*}

The lower bound in (\ref{eq:A}) relies on Lemma~\ref{lm:MinLength}.
Let $P=(x_0,\ldots,x_k)$ denote a path in $\G$ for which
\[
\sum_{i=1}^k \left\| x_{i-1}-x_i \right\| = 
d_G \left( x,y \right).
\]
From the construction of $\G$, each edge length
$\left\| x_{i-1}-x_i \right\| \leq \epsilon < s_0$
and it follows from the definition of minimum branch separation that $\ell_i = d_\M(x_{i-1},x_i) \leq \pi r_0$.  It then follows from Lemma~\ref{lm:MinLength} that $\left\| x_{i-1}-x_i \right\| \geq 2r_0 \sin ( \ell_i/2r_0)$.  Applying the trigonometric inequality $\sin u \geq u-u^3/6$ for $u \geq 0$, we obtain
\begin{equation}
\left\| x_{i-1}-x_i \right\| \geq 2r_0 \left[ \frac{\ell_i}{2r_0} - \frac{1}{6} \left( \frac{\ell_i}{2r_0} \right)^3 \right] =
\left( 1 - \frac{\ell_i^2}{24r_0^2} \right) \ell_i .
\label{eq:trig.strong}
\end{equation}
Applying the trigonometric inequality $\sin u \geq (2/\pi)u$ for $u \in [0,\pi/2]$, we obtain
\[
\frac{2}{\pi} \ell_i = 2r_0 \frac{2}{\pi} \frac{\ell_i}{2r_0}
\leq 2r_0 \sin \left( \frac{\ell_i}{2r_0} \right) \leq
\left\| x_{i-1}-x_i \right\| \leq \epsilon \leq
\frac{2}{\pi} r_0 \sqrt{24 \lambda},
\]
from which it follows that
\begin{equation}
\frac{\ell_i^2}{24r_0^2} \leq \lambda.
\label{eq:trig.weak}
\end{equation}
Combining (\ref{eq:trig.strong}) and (\ref{eq:trig.weak}) then yields $\left\| x_{i-1}-x_i \right\| \geq (1-\lambda) \ell_i$,
hence
\[
d_G \left( x,y \right) = 
\sum_{i=1}^k \left\| x_{i-1}-x_i \right\| \geq
(1-\lambda) \sum_{i=1}^k \ell_i \geq
(1-\lambda) d_\M(x,y).
\]
\hfill $\Box$

\medskip

Next we consider how to construct a finite set $V \subset \M$ that satisfies the $\delta$-sampling condition.  The following result is analogous to the Sampling Lemma in \cite{Bernstein&etal:2000}.
\begin{lemma}
For $r>0$, let
\[
B(m,r) = \left\{ x \in \M : d_\M(m,x) < r \right\}
\]
denote an open ball in the compact connected Riemannian manifold $\M$.
Let $\mu$ denote any probability measure on $(\M,\B)$ for which every $\mu(B(m,r))>0$, and suppose that $x_1,\ldots,x_n \iid \mu$.
Let $E_n$ denote the event that every $x \in \M$ lies within Riemannian distance $\delta$ of some $x_j$.  Then
$\lim_{n \rightarrow \infty} P(E_n)=1$.
\label{lm:sampling}
\end{lemma}

\subparagraph{Proof}
The collection of balls $\{ B(m,\delta/2) : m \in \M \}$ covers $\M$.  Because $\M$ is compact, we can extract a finite subcover of $\M$, say $B_1,\ldots,B_k$. 
Let $b = \min_i \mu(B_i)$ and note that $b>0$. 
The event $E_n$ obtains if each $B_i$ contains at least one $x_j$, which occurs with probability
\begin{eqnarray*}
\lefteqn{P \left( \mbox{every $B_i$ contains an $x_j$} \right)
 = 1- P \left( \mbox{some $B_i$ contains no $x_j$} \right)} & & \\ 
  & \geq & 1- \sum_{i=1}^k P \left( \mbox{$B_i$ contains no $x_j$} \right) 
  = 1- \sum_{i=1}^k \prod_{j=1}^n P \left( x_j \not\in B_i \right) \\
  & = & 1- \sum_{i=1}^k \prod_{j=1}^n \left(
  1 - \mu \left( B_i \right) \right) 
  \geq  1- \sum_{i=1}^k \prod_{j=1}^n \left(
  1 - b \right) 
  =  1-k \left(
  1 - b \right)^n,
\end{eqnarray*}
which tends to $1$ as $n \rightarrow \infty$.
\hfill $\Box$

\medskip

Combining Lemma~\ref{lm:sampling} and Theorem~\ref{thm:A}, we now establish that shortest path distances on $\epsilon$-neighborhood graphs converge uniformly to Riemannian distances.  The convergence analysis in \cite{Bernstein&etal:2000} also considers the more complicated case of $K$-nearest neighbor graphs, which are often preferred in practice.
\begin{theorem}
Let $\M \subset \Re^q$ be a compact connected Riemannian manifold and let $\mu$ be any probability measure on $(\M,\B)$ such that $\mu(B(m,r))>0$ for every $m \in \M$ and $r>0$.  Suppose that $x_1,x_2,\ldots \iid \mu$ and let $V_n = \{ x_1,\ldots,x_n \}$.  For $\epsilon>0$, let
$\G_{n,\epsilon}=(V_n,E_{n,\epsilon})$ denote the graph with vertex set $V_n$ and edges between $x_i$ and $x_j$ if and only if $\| x_i-x_j \| \leq \epsilon$.  Let $d_{n,\epsilon}$ denote shortest path distance on $\G_{n,\epsilon}$ with edge weights $\| x_i-x_j \|$.  
If $\delta_k \rightarrow 0$ as $k \rightarrow \infty$,
\[
 x_{a_k} \in V_{n_k} \mbox{ with } d_M \left( x_{a_k},m_a \right) < \delta_k^2/4, \mbox{ and } 
 x_{b_k} \in V_{n_k} \mbox{ with } d_M \left( x_{b_k},m_b \right) < \delta_k^2/4,
\]
then there exist sequences $n_k \rightarrow \infty$ and $\epsilon_k \rightarrow 0$
such that
\[
\sup_{m_a,m_b \in \M} \left|
d_{n_k,\epsilon_k} \left( x_{a_k},x_{b_k} \right) -
d_\M \left( m_a,m_b \right)
\right| \stackrel{P}{\rightarrow} 0.
\]
\label{thm:B}
\end{theorem}

\subparagraph{Proof}
Let $\alpha_k \rightarrow 0$ be a decreasing sequence of error probabilities.  Using Lemma~\ref{lm:sampling} with $\delta=\delta_k^2/4$, choose $n_k$ sufficiently large that $x_1,\ldots,x_{n_k} \iid \mu$ satisfies the $\delta$-sampling condition with probability at least $1-\alpha_k$.
Let $\bar{d} = sup_{x,y \in \M} d_\M(x,y)$ denote the diameter of $\M$ and note that $\bar{d} < \infty$ because $\M$ is compact.
Set $\lambda = \delta^{1/2}/\bar{d}$.

Suppose that the $\delta$-sampling condition obtains.  
Given $m_a,m_b \in \M$, choose $x_a,x_b \in V_{n_k}$ such that
$d_\M(x_i,m_i) < \delta$, $i=a,b$.  For $k$ large enough that $\delta_k< \min(3 r_0^2/\pi^2,1)$, choose $\epsilon_k$ so that
\[
\pi^2 \delta \leq \epsilon_k^2 \leq 6r_0^2\delta^{1/2},
\]
then apply Theorem~\ref{thm:A} to obtain
\begin{eqnarray*}
\lefteqn{\left| d_{n_k,\epsilon_k} \left( x_a,x_b \right) -
d_\M \left( m_a,m_b \right) \right|} & & \\
 & \leq &
\left| d_{n_k,\epsilon_k} \left( x_a,x_b \right) -
d_\M \left( x_a,x_b \right) \right| + 
d_\M \left( m_a,x_a \right) + d_\M \left( x_b,m_b \right) \\ & < &
\lambda d_\M \left( x_a,x_b \right) + 2\delta \leq 
\delta^{1/2} + 2\delta = \delta_k/2 + \delta_k^2/2 < \delta_k.
\end{eqnarray*}
\hfill $\Box$

\medskip

Assuming that one constructs an $\epsilon$-neighborhood graph to localize manifold structure,
Theorem~\ref{thm:B} provides theoretical justification for the first two steps of Isomap.  No convexity assumptions are required.  The pairwise Riemannian distances between points, hence the pairwise shortest path distances that approximate them, may not be Euclidean distances.  (Indeed, they are only Euclidean distances in very special cases.)  The third step of Isomap approximates the approximate Riemannian distances with Euclidean distances, constructing a plausible Euclidean representation of the manifold.
Apparent distortions such as appear in Section~\ref{annuli} are better understood as properties of the manifold than as failures of Isomap.

\section{Discussion}
\label{discuss}

Isomap combines two ideas: the approximation of Riemannian distances with shortest path distances on a graph that localizes manifold structure, and the approximation of shortest path distances with Euclidean distances by multidimensional scaling.  Isomap's novelty lies in the first idea, but its limitations may just as easily lie in the second.  From its introduction in 2000, Isomap has been presented, described and criticized as a technique that requires some form of convexity.  Re-examining the early literature on Isomap, it becomes apparent that the role of convexity was misunderstood.  Convexity is not required to ensure that shortest path distances approximate Riemannian distances, but a lack of convexity guarantees that the Riemannian distances are non-Euclidean.  Indeed, Riemannian distances are Euclidean only in very special cases.  The real challenge is not the problem of approximating Riemannian distances with shortest path distances, it is the problem of approximating non-Euclidean distances with Euclidean distances.  Isomap uses a standard methodology (multidimensional scaling) to address the latter problem, and that methodology does what can be done.  One should not blame Isomap if the manifold to be learned has a non-Euclidean geometry.

If a manifold's geometry is non-Euclidean, then one might object to  the entire project of constructing a Euclidean representation of the manifold's geodesic structure and prefer a completely different interpretation of what it means to learn a manifold.  Such an objection would invite one to inquire what one is trying to learn and why one is trying to learn it.  Dimension reduction may be an end in itself, but it may also be the prelude to a subsequent inference.  In fact, one can easily devise exploitation tasks for which the Euclidean approximation of geodesic structure is precisely what one wants to learn about the manifold.  

Consider, as in \cite{mwt:rdpg1}, the problem of drawing inferences about Fr\'{e}chet means on a Riemannian manifold $\M$, e.g., the $2$-sample problem of deciding whether or not the Fr\'{e}chet means of two populations are identical.  The {\em Fr\'{e}chet mean set}\/ of a probability measure $\mu$ on $\M$ is the set of all minimizers of the map $\mbox{\rm Fr} : \M \rightarrow \Re$ defined by
\[
\mbox{\rm Fr}(m) = \int_{\mathcal M} \left[
d_\M (m,x) \right]^2 \mu(dx).
\]
If a unique minimizer $\mu_{\mbox{\rm \tiny Fr}}$ exists, then it is the {\em Fr\'{e}chet mean}\/ of $\mu$.  The {\em sample Fr\'{e}chet mean}\/ of $x_1,\ldots,x_n \in \M$ is the Fr\'{e}chet mean of the empirical distribution of $x_1,\ldots,x_n$, a consistent estimator of $\mu_{\mbox{\rm \tiny Fr}}$, and a plausible test statistic is the Riemannian distance between the two sample Fr\'{e}chet means.
Unfortunately, these quantities are difficult to compute in practice.  In Euclidean space, however, a sample Fr\'{e}chet mean is simply the average of the $x_i$.  If one can construct a representation in which Riemannian distances are approximated by Euclidean distances, then the sample Fr\'{e}chet means can be approximated by averaging and the test statistic by the Euclidean distance between them.  Such a representation is precisely what Isomap constructs.  Of course, this representation is only an approximation and the power function of the resulting test will only converge to the power function of the desired test if the approximation is exact. 

To mitigate confusion, our exposition has glossed a subtle concern on which we now remark.  If $\M$ is a Riemannian manifold of dimension $p$, then discussions of Isomap invariably assume that the  desired representation will be constructed in a Euclidean space of dimension $d=p$.  In fact, what Isomap actually constructs is not the manifold itself but a Euclidean representation of the manifold's geodesic structure.  Typically, Riemannian distances are non-Euclidean, in which case the quality of the necessarily imperfect approximation of the Riemannian distances with Euclidean distances will depend on $d$.  It is perfectly reasonable to prefer $d>p$ dimensions in order to obtain a better representation of geodesic structure.

We admire Isomap for its elegant simplicity and believe that its virtues have been greatly underestimated.  Nevertheless, even if a Euclidean representation of geodesic structure is precisely what one desires, there are substantial research issues that remain to be addressed.  The convergence analysis of Isomap relies on densely sampling the manifold to be learned, something that is rarely possible in practice. 
Choosing a suitable value of the localization parameter for constructing the $\epsilon$-neighborhood or $K$-nearest neighborhood is a difficult problem that demands further investigation.  Both problems are exacerbated in the case of data that do not lie precisely on the manifold.  Ultimately, our objective is not so much to commend Isomap to practitioners as to insist that Isomap deserves further investigation by the manifold learning community.

\section*{Acknowledgments}
This work was partially supported by the Naval Engineering Education Consortium (NEEC), Office of Naval Research (ONR) Award Number N00174-19-1-0011.  The authors benefitted enormously from discussions with Lijiang Guo, Bruce Solomon, and Carey E. Priebe.

\bibliography{$HOME/lib/tex/stat,$HOME/lib/tex/mds,$HOME/lib/tex/math,$HOME/lib/tex/mwt,$HOME/lib/tex/cep,$HOME/lib/tex/bio,$HOME/lib/tex/proximity}

\end{document}